\documentclass[10pt,twocolumn,letterpaper]{article}

\usepackage[pagenumbers]{cvpr} 

\definecolor{cvprblue}{rgb}{0.21,0.49,0.74}
\usepackage[pagebackref,breaklinks,colorlinks,allcolors=cvprblue]{hyperref}
\usepackage[capitalize]{cleveref}
\usepackage{graphicx}
\usepackage{float}
\usepackage{adjustbox}
\usepackage[all]{hypcap}

\title{Prune-Then-Plan: Step-Level Calibration for Stable Frontier Exploration in Embodied Question Answering}

\author{
Noah Frahm
~~~~
Prakrut Patel
~~~~
Yue Zhang
~~~~
Shoubin Yu
~~~~
Mohit Bansal
~~~~
Roni Sengupta\\
University of North Carolina at Chapel Hill
}

\begin{document}

\twocolumn[{%
  \renewcommand\twocolumn[1][]{#1}%
  \maketitle
  \begin{center}
    \includegraphics[width=\linewidth]{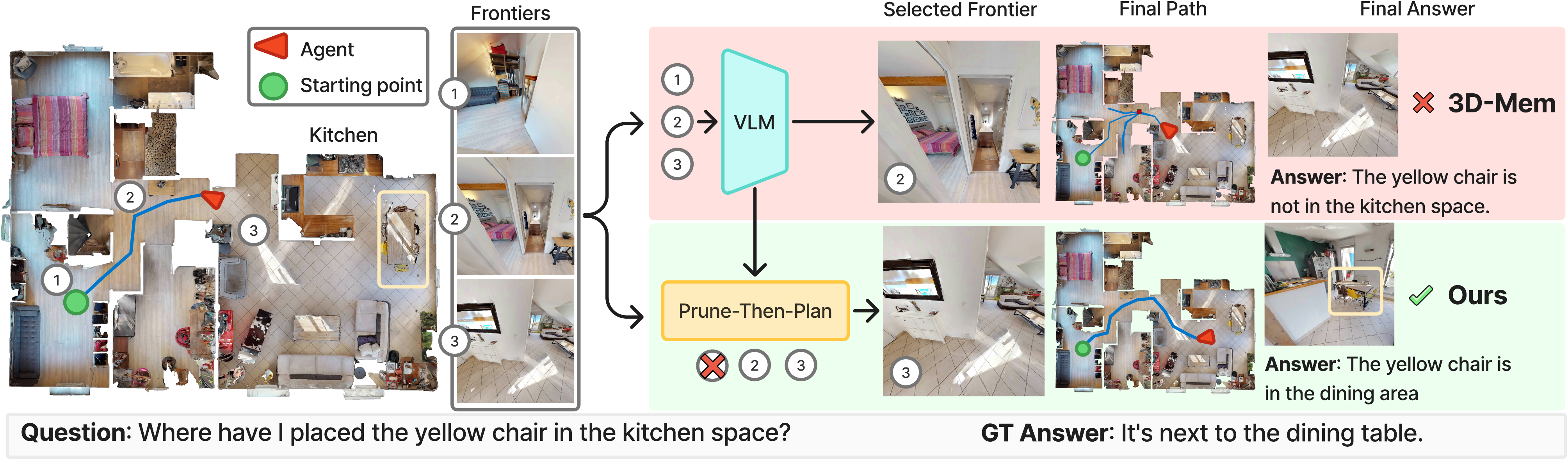}
    \captionof{figure}{Our method stabilizes VLM-guided exploration by calibrating frontier choices at the step level. Rather than letting the VLM directly pick the `best’ frontier (as in 3D-Mem), we use the VLM only to flag and reject frontiers that are likely uninformative. Once pruning is complete, a coverage-based planner selects the next frontier from the remaining candidates. For instance, although the VLM might favor frontier \textcircled{2}, our calibration step rejects frontier \textcircled{1} as a bad option (since it leads the agent away from the kitchen) and then selects the closest viable frontier \textcircled{3}, enabling the agent to reach the correct, visually grounded answer more quickly. This separation of semantic pruning and coverage-based planning provides a principled balance, allowing the agent to explore efficiently while maintaining strong semantic relevance.
    }
    \label{fig:teaser}
  \end{center}
}]

\begin{abstract}

Large vision-language models (VLMs) have improved embodied question answering (EQA) agents by providing strong semantic priors for open-vocabulary reasoning. However, when used directly for step-level exploration, VLMs often exhibit frontier oscillations, unstable back-and-forth movements caused by overconfidence and miscalibration, leading to inefficient navigation and degraded answer quality. We propose Prune-Then-Plan, a simple and effective framework that stabilizes exploration through step-level calibration. Instead of trusting raw VLM scores, our method prunes implausible frontier choices using a Holm-Bonferroni inspired pruning procedure and then delegates final decisions to a coverage-based planner. This separation converts overconfident predictions into conservative, interpretable actions by relying on human-level judgments to calibrate the step-level behavior of VLMs. Integrated into the 3D-Mem EQA framework, our approach achieves relative improvements of up to 49\% and 33\% in visually grounded SPL and LLM-Match metrics respectively over baselines. Overall, our method achieves better scene coverage under equal exploration budgets on both OpenEQA and EXPRESS-Bench datasets. We provide additional visuals of results on our
\href{https://noahfrahm.github.io/Prune-Then-Plan-project-page/}{\textbf{Project Page}}.

\end{abstract}    
\vspace{-1em}
\section{Introduction}
\label{sec:intro}

Embodied Question Answering (EQA) agents aim to navigate a 3D environment, gather visual observations, and answer natural language questions grounded in the scene. Recent advances in large vision-language models (VLMs) and large language models (LLMs) have substantially improved the open-vocabulary reasoning capabilities of these agents. By leveraging pretrained multimodal priors, VLM-powered EQA systems can now ground observations, maintain semantic memories, and reason about high-level tasks~\cite{hong20233dllminjecting3dworld, yang20253dmem3dscenememory, ren2024exploreconfidentefficientexploration, saxena2025grapheqausing3dsemantic, hu20253dllmmemlongtermspatialtemporalmemory, sakamoto2024mapbasedmodularapproachzeroshot, zhao2025cityeqahierarchicalllmagent, EXPRESSBench, zhang2024vision, huang2024embodied}. 

Despite these advances, directly delegating step-level exploration to VLMs remains brittle. Agents cannot feasibly pass complete scene context or full action histories at every step due to token and compute limits, and long prompts are known to degrade reasoning and retrieval quality~\cite{du2025contextlengthhurtsllm, zhou2024rethinkingvisualdependencylongcontext, liu2023lostmiddlelanguagemodels, tian2025identifyingmitigatingpositionbias}. 
Moreover, recent calibration studies have shown that VLMs are frequently overconfident and poorly calibrated~\cite{ren2023robotsaskhelpuncertainty, ren2024exploreconfidentefficientexploration, EXPRESSBench, vo2025visionlanguagemodelsbiased, groot2024overconfidencekeyverbalizeduncertainty, zollo2025confidencecalibrationvisionlanguageactionmodels, li2023evaluatingobjecthallucinationlarge, kostumov2024uncertaintyawareevaluationvisionlanguagemodels}, leading to unreliable confidence estimates in downstream embodied tasks.

Prior works~\cite{ren2024exploreconfidentefficientexploration, EXPRESSBench, ren2023robotsaskhelpuncertainty} often rely on `frontier' points to perform exploration, where frontier refers to the reachable points at the boundary between the explored and unexplored regions of the environment. 
These approaches then address the unreliability of VLMs by calibrating the full episode-level behavior, for instance, deciding when to terminate exploration once the agent is confident enough to answer. However, these approaches overlook a critical issue when VLMs are used directly for step-level exploration: miscalibration of VLM during exploration can lead to unstable and oscillatory behavior. This leads to agents often taking significantly longer to answer correctly, answering incorrectly due to exploration leading to the wrong direction, or complete failure to answer due to a limit in exploration steps, as shown in~\Cref{fig:path_compare}. Existing systems attempt to mitigate this indirectly, e.g., by embedding VLM guidance in semantic maps, scene graphs, or hierarchical planners~\cite{ren2024exploreconfidentefficientexploration, saxena2025grapheqausing3dsemantic, EXPRESSBench}, but none directly address step-level instability caused by overconfident VLM decisions.

In this work, we focus on solving the unstable and inefficient exploration behavior of EQA agents, which often leads to wrong answers or timeouts. Our key idea is to improve exploration behavior by explicitly calibrating the step-level exploration decisions of the VLM-guided EQA agent. We introduce a simple yet effective `prune-then-plan' framework that explicitly calibrates VLM confidence to the step-level of the exploration. Rather than trusting raw VLM scores to select a single frontier, our method first statistically rejects implausible frontiers based on calibrated confidence, and then delegates the final selection to a coverage planner that optimizes interpretable heuristics such as distance-to-frontier. We calibrate the VLM's confidence in predicting the effectiveness of frontier snapshots at each step of the exploration using human-level judgments. This separation of pruning and planning turns the VLM’s overconfidence into a conservative, stable exploration strategy. We adopt the Holm–Bonferroni multiple hypothesis testing procedure~\cite{MR538597} as a structured rejection rule over frontier hypotheses and treat $\alpha$ as a behavioral hyperparameter that controls how aggressively frontiers are pruned, rather than as a formal error-control level. When the VLM is confident, pruning collapses to a single choice; when it is uncertain, a larger frontier set remains, allowing the planner to choose a safe next move. We observe that this leads to significantly better exploration behavior of the EQA agent, reducing unnecessary frontier oscillations, resulting in more efficient and correct answers, as shown in ~\Cref{fig:teaser} and ~\Cref{fig:path_compare}.

We then integrate our `Prune-Then-Plan' module into the 3D-Mem EQA framework~\cite{yang20253dmem3dscenememory} with snapshot memory and frontier-based exploration. On OpenEQA and EXPRESS-Bench~\cite{majumdar2023openeqa, EXPRESSBench}, Our method improves SPL and LLM-Match by up to 49\% and 29\% over baselines, and achieves better scene coverage under equal exploration budgets. Our key contributions are: 
\begin{itemize}
    \item \textit{Step-level calibration for exploration:} We propose our Prune-Then-Plan scheme that uses a VLM solely for frontier rejection, delegating the final move to a planner for stable and interpretable behavior. 
    \item \textit{Tunable Frontier pruning:} We introduce a per-step Holm–Bonferroni~\cite{MR538597} pruning rule controlled by a tuning parameter $\alpha$, enabling trade-offs between exploration stability, coverage, and answer quality. 
    \item \textit{System-level integration and evaluation:} We integrate our approach into an image-centric EQA pipeline with minimal overhead and show consistent gains in path efficiency, overall correctness, and exploration behavior.
\end{itemize}

\section{Related Works}
\label{sec:related_works}

\noindent \textbf{VLMs for Embodied Reasoning.}
Embodied Question Answering (EQA)~\cite{das2017embodiedquestionanswering} tasks an agent with answering questions about an unknown scene where it must actively explore a 3D environment to gather the information needed to answer. Early EQA formulations, such as episodic-memory EQA (EM-EQA), rely on pre-recorded trajectories or known environments, limiting the agent’s ability to adapt the exploration to the question. In contrast, Active Embodied Question Answering (A-EQA)~\cite{majumdar2023openeqa} places the agent in an unknown environment and allows it to freely explore the scene to correctly answer the question. This setting is more challenging since the agent cannot rely on a pre-recorded trajectory that is guaranteed to have the information it needs, making exploration policy, memory construction, and retrieval critical for good performance.

Recent progress in Vision-Language Models (VLMs) and Large Language Models (LLMs) has significantly impacted EQA research. VLMs such as BLIP, CLIP, and Flamingo~\cite{li2022blipbootstrappinglanguageimagepretraining, li2023blip2bootstrappinglanguageimagepretraining, radford2021learningtransferablevisualmodels, alayrac2022flamingovisuallanguagemodel} provide strong priors learned from large-scale paired data. These models contribute general reasoning ability and compositional semantics, enabling agents to interpret goals, explain decisions, and generate textual justifications. However, these models are often poorly calibrated leading to overconfident answers~\cite{ren2023robotsaskhelpuncertainty, ren2024exploreconfidentefficientexploration,
EXPRESSBench, vo2025visionlanguagemodelsbiased, groot2024overconfidencekeyverbalizeduncertainty, zollo2025confidencecalibrationvisionlanguageactionmodels, li2023evaluatingobjecthallucinationlarge, kostumov2024uncertaintyawareevaluationvisionlanguagemodels}. Another problem is that VLMs and LLMs are temporally stateless, relying on the prompt to provide sufficient context about what the agent state and action history. Larger prompts come with the risk of text bias~\cite{du2025contextlengthhurtsllm, zhou2024rethinkingvisualdependencylongcontext, liu2023lostmiddlelanguagemodels, tian2025identifyingmitigatingpositionbias, deng2025wordsvisionvisionlanguagemodels, zheng2025unveilingintrinsictextbias} which can lead to poor visual reasoning and hurt the overall EQA performance. Consequently, several works explore grounding LLM or VLM reasoning in episodic memory, spatial priors, or exploration history to improve reliability~\cite{yang20253dmem3dscenememory, saxena2025grapheqausing3dsemantic, ren2024exploreconfidentefficientexploration, EXPRESSBench}.

\noindent \textbf{Representation Learning for EQA.} A key design choice in EQA systems is the world representation, how to effectively store what the agent has seen for efficient retrieval and reasoning. Common approaches include scene graphs which encode objects and their spatial/semantic relations~\cite{pmlr-v164-li22e, Werby_2024, saxena2025grapheqausing3dsemantic}, semantic maps which track semantically-relevant regions~\cite{ren2024exploreconfidentefficientexploration, EXPRESSBench}, and image-centric memories structured by graphs that retain key views for re-localization and VLM reasoning~\cite{saxena2025grapheqausing3dsemantic, yang20253dmem3dscenememory}. World representations help in information retrieval but can also help EQA methods guide exploration. Exploration efficiency can strongly affect downstream performance, if the agent is able to gather more relevant information in a shorter amount of time, it can provide better and faster answers. The most widely used paradigm for exploration is frontier-based exploration~\cite{frontierexploration}, where frontiers denote the boundaries between explored and unexplored regions. Correctly identifying and selecting frontiers at each step can maximize information gain and improve navigation efficiency. Recent methods couple semantic maps with frontier reasoning, prioritizing focus on more semantically aligned frontiers~\cite{ren2024exploreconfidentefficientexploration, yokoyama2024vlfm, EXPRESSBench}. 3D-Mem~\cite{yang20253dmem3dscenememory} introduces the idea of frontier snapshot reasoning, allowing a VLM to control exploration by visually comparing and choosing from a set of candidate frontier images. Defining robust frontier selection for VLM controlled systems remains challenging, motivating integration with calibration-aware decision strategies.

\begin{figure*}[ht]
  \centering
  \includegraphics[width=\linewidth]{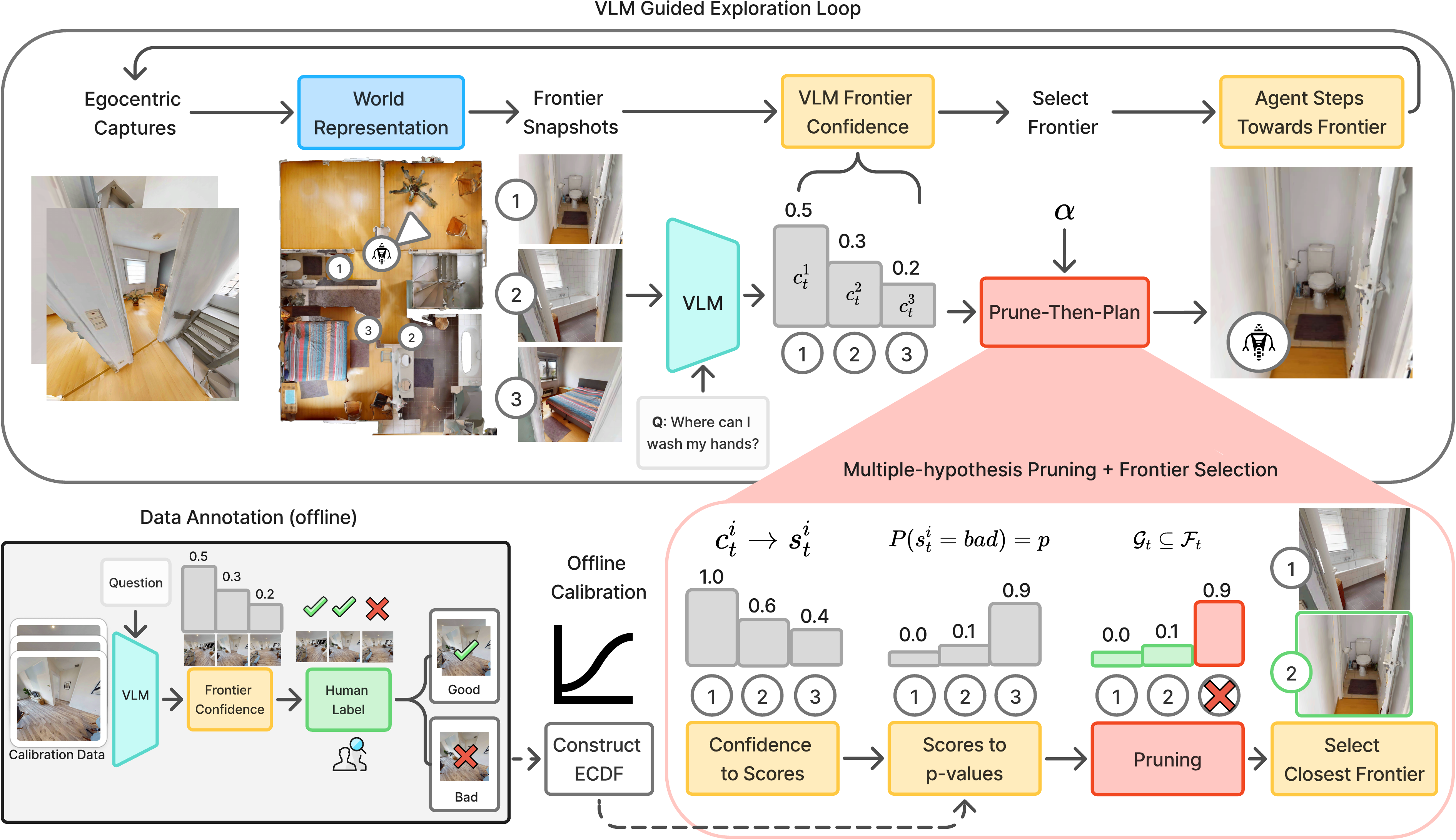}
   \caption{
   The agent traverses the scene and passes egocentric captures to the 3D-Mem world representation to update scene memory and compute frontiers. We subsequently query the VLM to assess its confidence in each frontier’s potential to move the agent closer to a correct answer. The resulting confidences are converted into step-normalized scores and then into p-values via our empirical cumulative distribution function to support pruning. Finally, we employ multiple hypothesis testing to detect and prune bad frontiers where $\alpha$ controls the aggressiveness of frontier pruning (larger $\alpha$ means more frontiers retained). The agent then proceeds towards the nearest surviving frontier and repeats the process.
   }
   \vspace{-1 em}
   \label{fig:inference_flow_and_prune}
\end{figure*}

\vspace{-0.5 em}
\section{Method}
\label{sec:method}

We stabilize VLM-guided exploration in EQA by focusing on step-level exploration choices, pruning the frontier snapshot options at each step. The VLM is used solely to discard frontiers that are unlikely to be informative, after which a coverage-based planner selects the best remaining frontier. This exploration strategy is integrated into the 3D-Mem~\cite{yang20253dmem3dscenememory} framework.
We begin by introducing the problem (\cref{method:setting}) and summarizing our approach (\cref{method:inference_flow}). We then provide a description of our approach in steps: (i) VLM-based frontier scoring (\cref{method:vlm_scoring}), (ii) human-labeled calibration and empirical cumulative distribution function based p-value mapping (\cref{method:calibration}), (iii) step-wise multiple-hypothesis testing and pruning (\cref{method:holm_bonferroni_testing}), and (iv) coverage-based decision making (\cref{method:coverage_planner}).

\subsection{Problem Setup} \label{method:setting}
Given a question about an unknown 3D scene, the agent must gather enough evidence to answer within a fixed exploration budget. Our agent is built directly on the 3D-Mem framework~\cite{yang20253dmem3dscenememory}, relying on its ability to maintain an image-centric scene memory and compute navigable frontiers. At each step $t$, the agent is presented with $K_t$ frontiers,

\vspace{-0.5 em}
\[
\mathcal{F}_t = \{f_t^1, f_t^2, \dots, f_t^{K_t}\},
\]
from which a single frontier must be selected. The goal is to select frontiers in a way that improves the agents ability to answer correctly. Directly relying on the VLM to make the final choice can result in back-and-forth oscillations across steps as shown in~\Cref{fig:path_compare}. 
Instead, we convert VLM confidences into calibrated scores and prune frontiers that are likely \emph{bad}.

We define a frontier as \emph{bad} if it leads to an area that is clearly not relevant to the current question, or does not meaningfully advance exploration of the scene (continues into a dead end or into a part of a room that was already observed and is unlikely to reveal new objects). For example, if the question asks about the kitchen, a frontier that takes the agent into a bathroom is bad; likewise, if the agent would move further along a corridor or deeper into a room where no new doors are visible, we also consider that frontier to be bad. For instance, as illustrated in~\Cref{fig:inference_flow_and_prune}, frontier 3 depicts a bedroom scene, which is clearly uninformative for addressing the question "Where can I wash my hands?" and our method prunes this frontier.

\subsection{Method Overview}
\label{method:inference_flow}

We provide an overview of our EQA pipeline in~\Cref{fig:inference_flow_and_prune} and detail the steps in this section. At each step $t$:

\noindent\textbf{Frontier confidence.} The VLM is queried with the set of frontier snapshots defining the agent’s possible choices. The corresponding confidence scores are extracted and normalized to produce a step-wise representation of the underlying confidence distribution over frontiers.

\noindent\textbf{p-value mapping.} We then transform the step’s frontier confidences into scores and map them to p-values using an empirical cumulative distribution function (ECDF) fitted on a held-out calibration set of bad frontiers. High p-values indicate scores that are typical for bad frontiers and are therefore candidates for pruning, while low p-values indicate scores that are unusual for bad frontiers, therefore more likely to be worth exploring. In our experiments, we use an ECDF calibrated on OpenEQA and observe performance gains which also transfers to EXPRESS-Bench, suggesting that the learned bad-frontier score distribution is reasonably stable across these settings.

\noindent\textbf{Multiple hypothesis testing.} We apply the Holm–Bonferroni~\cite{MR538597} style step-down rule over all $K_t$ frontier p-values in the current step to help us identify and prune bad choices from $\mathcal{F}_t$. We treat the set of frontiers as $K_t$ simultaneous tests of 

\vspace{-0.75em}
\[
H_0^i : \text{frontier } f_t^i \text{ is bad}
\]
\vspace{-1.0em}

In the classical setting, Holm–Bonferroni controls the family-wise error rate at level $\alpha$.
In our step-level setting we do \emph{not} interpret $\alpha$ as a formal error-control level; instead we reuse Holm–Bonferroni’s ordering-and-thresholding scheme as a structured rejection rule.
We treat $\alpha$ as a behavioral hyperparameter that controls how aggressively we prune frontiers. The frontiers that survive pruning form the accepted set $\mathcal{G}_t \subseteq \mathcal{F}_t$.

\noindent\textbf{Coverage-based selection.} Among $\mathcal{G}_t$, we choose the frontier with minimum travel cost, measured as the geodesic distance from the agent’s current position in the scene, i.e., a closest-frontier rule. This separates the VLM’s role in semantic pruning from the planner’s role in exploration efficiency and provides a simple, temporally stable objective (moving toward a chosen frontier tends to keep it closest) that discourages oscillation and is travel cost aware.

\subsection{VLM-Based Frontier Scoring}
\label{method:vlm_scoring}

\paragraph{Confidence extraction.} 
The original 3D-Mem prompt interleaves memory and frontier snapshots in a single prompt and passes it to the VLM. To extract frontier-specific confidence values, we adapt this prompt so that each frontier image is tagged with a unique letter token. This enables us to trace each output token back to its corresponding input image and compute token-level probabilities as confidence scores. 
Let $z_t^i$ denote the next-token logit associated with frontier $i$ at step $t$. We apply a softmax over all frontier logits at that step to obtain confidences $c_t^i$. We rely on token-level probabilities rather than requesting explicit confidence scores because such self-reported scores are known to be poorly calibrated~\cite{groot2024overconfidencekeyverbalizeduncertainty}. Token-level probabilities offer a well-established and reliable confidence signal~\cite{geng2024surveyconfidenceestimationcalibration, petryk2023simpletokenlevelconfidenceimproves, li2024referencefreehallucinationdetectionlarge, jiang2021knowlanguagemodelsknow} and avoid complex output parsing. 

\paragraph{Step-wise scoring.}
\label{method:step_scoring}
To obtain a calibration model that can be reused across episodes, we require a frontier score that is comparable across steps. If we were to fit on raw confidence values, the resulting ECDF would struggle to effectively map the underlying distribution of bad frontiers and would lose discriminative power. We therefore convert frontier confidences into \emph{relative} step-wise scores. For a step $t$ with frontier set $\{f_t^1,\dots,f_t^{K_t}\}$ and corresponding confidences $\{c_t^1,\dots,c_t^{K_t}\}$, we define
\vspace{-0.5em}
\[
s_t^i = \frac{c_t^i}{\max_j c_t^j} \in (0,1].
\]
\vspace{-0.5em}

The highest-confidence frontier at step $t$ attains $s_t^i = 1$ and serves as a reference, while all other are expressed as a fraction of this maximum. This normalization largely removes arbitrary multiplicative scale within each step and makes scores less sensitive to the absolute magnitude of the VLM’s confidence. Pooling $\{s_t^i\}$ across episodes then allows us to build an ECDF that reflects the typical relative scores of bad frontiers on a consistent, comparable scale.

\subsection{Calibration and ECDF Construction}
\label{method:calibration}

\paragraph{Human-labeled calibration data.}
To estimate how \emph{bad} frontiers are typically scored, we first build a calibration set of such scores from a held-out split of OpenEQA A-EQA scenes~\cite{majumdar2023openeqa}. We run the baseline agent and, at each step, record the current frontier set, the VLM-derived frontier confidences, and the step-normalized scores $s_t^i$ (\cref{method:step_scoring}). We then collect human labels indicating whether each frontier $f_t^i$ is \emph{bad} under the definition in \cref{method:setting} (question-irrelevant or offering negligible marginal information gain given the visible geometry). Two annotators independently label data which contains the question, the candidate frontier snapshots, and the occupancy grid, but not future frames or memory. Their task is to identify all frontiers that should be considered bad at the current step. In total we end up with approximately 5,500 labeled frontiers. To avoid producing a degenerate point mass at $s=1$, we exclude per-step argmax frontiers from the calibration pool, even when labeled as bad. Because the calibration set serves only to characterize the score distribution of clearly undesirable frontiers, we expect that moderate annotation noise may be less harmful than in tasks that require precise decision boundaries and further explore this in \cref{ablations:alpha_ablation}. 

\paragraph{ECDF construction and p-value mapping.}
From all bad-labeled frontiers in the calibration split we obtain a set of scores $\mathcal{S}_{\text{bad}} = \{ s^{(1)}, s^{(2)}, \dots, s^{(N)} \}$ and fit an ECDF

\vspace{-0.5em}
\[
\hat{F}_{\text{bad}}(x) = \frac{1}{N} \sum_{n=1}^N \mathbf{1}\{ s^{(n)} \le x \},
\]
which estimates the probability that a bad frontier attains a score at most $x$. At inference time, a step-normalized score $s_t^i$ is mapped to a right-tail p-value using the complementary ECDF with finite-sample smoothing:

\vspace{-0.5em}
\[
p_t^i = \frac{1 + N \bigl(1 - \hat{F}_{\text{bad}}(s_t^i)\bigr)}{1 + N}.
\]
\vspace{-0.5em}

Lower scores yield larger p-values 
which are potential evidence of bad frontiers, while higher scores yield smaller p-values (evidence against being bad). These p-values are then used in our step-wise multiple-hypothesis testing and pruning procedure (\cref{method:holm_bonferroni_testing}) to remove bad frontiers.

\subsection{Holm-Bonferroni Frontier Pruning}
\label{method:holm_bonferroni_testing}

At step $t$ we have $K_t$ p-values $\{p_t^i\}$. We sort them in ascending order $p_t^{(1)} \le \dots \le p_t^{(K_t)}$ with corresponding frontiers $f_t^{(1)}, \dots, f_t^{(K_t)}$. Given a target filtering strictness $\alpha$ we follow a Holm-Bonferroni method~\cite{MR538597} inspired step-down rule and find the largest index $m$ such that

\vspace{-0.5em}
\[
p_t^{(j)} \le \frac{\alpha}{K_t - j + 1} \quad \text{for all } j \le m,
\]
and reject our null-hypotheses $H_0^{(1)}, \dots, H_0^{(m)}$. 
Recall that $H_0^i : \text{``frontier } f_t^i \text{ is bad''}$. Intuitively, a small $p_t^i$ means that $f_t^i$ scored unusually high compared to typical bad frontiers, so we keep it. A large $p_t^i$ means the score is consistent with what we often see for bad frontiers, so we prune it. The kept frontiers form the accepted set

\vspace{-0.5 em}
\[
\mathcal{G}_t = \{ f_t^{(1)}, \dots, f_t^{(m)} \}.
\]

In practice, $\alpha$, which is typically a bound on the family-wise-error-rate, is treated as a \emph{behavioral} hyperparameter: smaller $\alpha$ yields a smaller frontier set (more aggressive pruning), while larger $\alpha$ retains more frontiers and defers more work to the coverage planner. We report results on our tuning set across several $\alpha$ values in \cref{ablations:alpha_ablation}.

\subsection{Coverage-Based Action Selection}
\label{method:coverage_planner}

After pruning bad frontiers, we ignore the VLM preference and treat all $f \in \mathcal{G}_t$ as equally viable, letting the navigation module choose closest frontier relative to the agents current position $\text{agent}_t$:

\vspace{-0.5em}
\[
f_t^\star = \arg\min_{f \in \mathcal{G}_t} \text{dist}(\text{agent}_t, f),
\]
where $\text{dist}(\cdot, \cdot)$ is the shortest geodesic-path distance on the agent’s current occupancy grid. This choice has two desirable properties:

1. \textbf{Error localization.} If a bad frontier survives pruning and is also the closest option, the agent will take a step towards it. This incurs only a short detour and produces additional observations from that region, which often helps the VLM assign lower confidence to that direction in later steps and higher confidence to more relevant frontiers.

2. \textbf{Coverage stability.} If a bad frontier survives pruning but is not the closest option, the planner will not visit it in that step. Instead, the agent advances toward another accepted frontier, which adds new observations. These additional observations provide the VLM with a richer context on the scene, so in later steps it can often better separate truly relevant frontiers from bad ones.

We re-run the entire procedure at every step $t$ so that new captures can update 3D-Mem's~\cite{yang20253dmem3dscenememory} internal occupancy map and compute updated frontiers to guide navigation.

\section{Experiments}
\label{sec:Experiments}

\begin{table*}[t]
  \centering
  \caption{Comparison of different EQA exploration methods on OpenEQA and Express-Bench. Metrics include visually grounded efficiency (SPL) and answer quality (LLM-Match/EAC), scene coverage (Coverage AUC), and path smoothness (Curvature).
  }
  \label{tab:openeqa_express_overall}
  \begin{adjustbox}{max width=\textwidth}
      \begin{tabular}{lcccccccc}
        \toprule
        & \multicolumn{4}{c}{OpenEQA} & \multicolumn{4}{c}{Express-Bench} \\
        \cmidrule(lr){2-5} \cmidrule(lr){6-9}
        Method
          & SPL $\uparrow$
          & LLM-Match/EAC $\uparrow$ 
          & Cov AUC $\uparrow$ 
          & Curvature $\downarrow$ 
          & SPL $\uparrow$ 
          & LLM-Match/EAC $\uparrow$ 
          & Cov AUC $\uparrow$ 
          & Curvature $\downarrow$  \\
        \midrule
        Fine-EQA~\cite{EXPRESSBench}
          & 19.5 & 25.7 & 0.14 & 84.4
          & 20.0 & 38.0 & 0.13 & 81.0 \\
        3D-Mem~\cite{yang20253dmem3dscenememory}
          & 24.7 & 29.2 & 0.14 & 70.4
          & 25.6 & 37.8 & 0.14 & 66.7 \\
        Ours
          & \textbf{29.1} & \textbf{34.3} & \textbf{0.19} & \textbf{45.9}
          & \textbf{25.7} & \textbf{39.5} & \textbf{0.19} & \textbf{45.8} \\
        \bottomrule
      \end{tabular}
  \end{adjustbox}
  \vspace{-1em}
\end{table*}

\begin{figure*}
  \centering
  \includegraphics[width=0.9\linewidth]{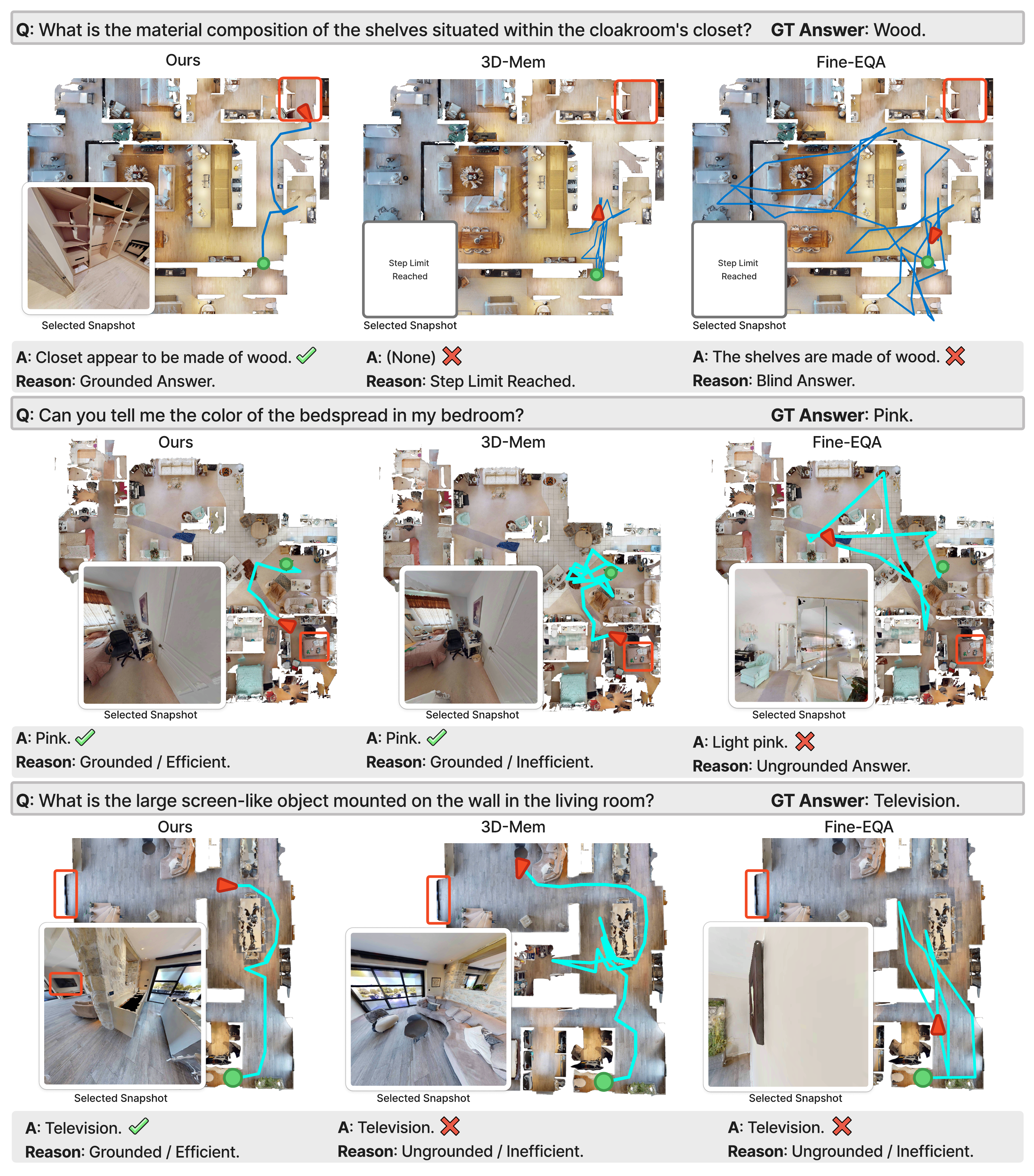}
   \caption{Baseline methods with inefficient exploration strategies suffer from extreme failure cases and can directly hurt answer quality. As seen in the top row, the efficiency of our method allows it to reach supporting visual evidence within the step budget where-as 3D-Mem and Fine-EQA both fail because their inefficiencies prevent them from ever reaching the region of interest. In the second row we see that even when 3D-Mem answers correctly it can take much longer to do so. In the bottom row we see a failure case where the baseline methods both answer but do not provide answers visually grounded in the selected snapshots.}
   \label{fig:path_compare}
   \vspace{-1em}
\end{figure*}

\subsection{Evaluation Framework}

\noindent \textbf{Dataset.}
\textit{A-EQA of OpenEQA}~\cite{majumdar2023openeqa} is an EQA dataset that spans various question categories on scenes from the HM3D~\cite{ramakrishnan2021habitatmatterport3ddatasethm3d} dataset. We use the A-EQA portion of OpenEQA~\cite{majumdar2023openeqa}, which provides scene question pairs along with agent starting positions. For our evaluation, we use the same 184 A-EQA question subset used by 3D-Mem~\cite{yang20253dmem3dscenememory}. The remaining 348 A-EQA questions are used to generate calibration data to construct the ECDF for frontier filtering, as detailed in \cref{method:calibration}.

\textit{Express-Bench}~\cite{EXPRESSBench} is a recent large scale EQA dataset that focuses on ensuring answers are well grounded in exploration. This dataset includes more than four times as many A-EQA style questions as OpenEQA and presents the new exploration answer consistency (EAC) metric for evaluating with visual grounding in mind. We reserve a 252 question subset for tuning filtering strictness $\alpha$ and show evaluation results on the remaining 1787 questions in \cref{sec:Experiments}.

\noindent \textbf{Baselines.}
We compare our method with two state-of-the-art embodied 3D-EQA approaches, 3D-Mem~\cite{yang20253dmem3dscenememory} and Fine-EQA~\cite{EXPRESSBench}. 3D-Mem relies on uncalibrated VLM responses to make direct step level exploration decisions based on retrieved memory and frontier snapshots from it's world representation. At each step 3D-Mem chooses from all frontiers with no filtering. Our method uses the 3D-Mem world representation, but we directly modify the exploration strategy, no longer allowing a VLM to directly select the next frontier. Fine-EQA builds a question-specific semantic map using VLM scores during exploration and uses this to help select relevant exploration points in the scene and to determine when to stop exploration, without directly using the VLM for step level frontier choices. For all baselines and calibration we run the agent using the open source Qwen 2.5-VL 32B model~\cite{qwen2025qwen25technicalreport} and set a per-scene step limit of 50. For visually grounded scoring (LLM-Match/EAC and SPL), we use the stronger Qwen 3-VL 30B model~\cite{qwen3technicalreport}.

\noindent \textbf{Metrics.}
\textit{(1) EQA Metrics.} OpenEQA introduces LLM-Match~\cite{majumdar2023openeqa} which uses an LLM to score free-form answers from (1–5) and also uses that score to replace the binary success term in SPL~\cite{anderson2018evaluationembodiednavigationagents}, providing an answer-aware path-efficiency measure. However, LLM-Match does not ensure that answers are visually grounded in an agent’s observations. EXPRESS-Bench addresses this with the Exploration–Answer Consistency (EAC) metric~\cite{EXPRESSBench}, combining LLM-Match with a VLM-based visual grounding score to check whether the answer is supported by seen imagery. The grounded score replaces LLM-Match in SPL, setting efficiency to zero when answers lack visual justification. We adopt EXPRESS-Bench metrics for stricter, fairer evaluation across datasets.

\noindent\textit{(2) Exploration-Behavior Metrics.} 
While EAC~\cite{EXPRESSBench} reflects final answer quality and path efficiency, it overlooks step-level exploration behavior. To capture this, we measure path curvature as the mean turning angle between consecutive steps as a proxy for frontier oscillations, since excessive backtracking produces high curvature and less smooth trajectories (~\Cref{fig:path_compare}). We also evaluate scene coverage vs. steps and report its area under the curve (AUC), which reflects how efficiently methods convert steps into new observations. Lower coverage speed or frequent oscillations naturally yield smaller AUC values.

\subsection{Results}
\label{ssec:results}

\noindent \textbf{Question Answer Quality.}
We compare our method against competitive EQA baselines, 3D-Mem~\cite{yang20253dmem3dscenememory} and Fine-EQA~\cite{EXPRESSBench} on both OpenEQA and EXPRESS-Bench datasets. As shown in \cref{tab:openeqa_express_overall}, our approach consistently outperforms both 3D-Mem and Fine-EQA across all metrics, including SPL and LLM-Match/EAC. Our method delivers significant improvements over Fine-EQA, boosting SPL and LLM-Match metrics by 49\% / 33\% on OpenEQA and 29\% / 4\% on EXPRESS-Bench. Similarly, when compared to 3D-Mem, our approach improves SPL and LLM-Match metrics, with relative improvements of 18\% / 21\% on OpenEQA and 0.3\% / 4\% on EXPRESS-Bench. Our improvements over Fine-EQA are significantly higher in SPL than in LLM-match since Fine-EQA lacks a scene memory and relies solely on egocentric views to decide when to answer, leading to more exploration when the current view is not sufficiently informative, reducing path efficiency. The smaller performance gap on EXPRESS-Bench between Ours and 3D-Mem may stem from the fact that short-horizon tasks dominate and many routes where 3D-Mem and our method overlap, diluting gains observed on more challenging, longer-horizon questions, as shown in visual examples in ~\Cref{fig:path_compare} and more in supplementary.

\begin{figure}
  \centering
  \includegraphics[clip, width=\linewidth]{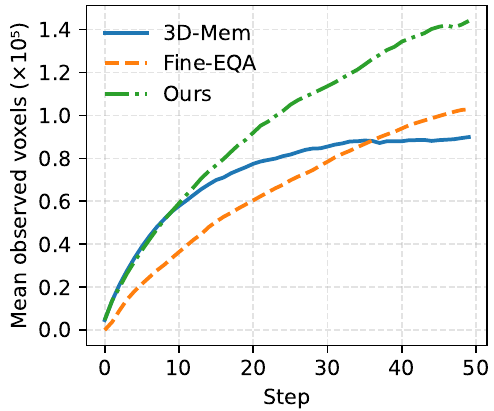}
   \caption{Mean observed voxels at step x for Express-Bench, obtained by averaging the number of voxels seen by all questions that reach that step.}
   \vspace{-0.5em}
   \label{fig:coverage_curves}
\end{figure}

\noindent \textbf{Exploration Behavior.}
As shown in \cref{tab:openeqa_express_overall}, our method surpasses all baselines in both coverage AUC and path curvature. On EXPRESS-Bench, it reduces curvature by 43\% over Fine-EQA and 31\% over 3D-Mem, and by 46\% and 35\% on OpenEQA, respectively. This is because our method significantly reduces oscillatory motion in frontier selection, which also leads to a decrease in average step-to-step turning angles, hence lower curvature, as shown in~\Cref{fig:path_compare}. We also observe notable gains in coverage AUC across both datasets, indicating that our method converts steps into new observations more efficiently. As shown in ~\Cref{fig:coverage_curves}, 3D-Mem initially matches our coverage rate but plateaus in later steps, reflecting limited long-horizon exploration. Fine-EQA starts with lower coverage but maintains steadier growth; however, our method consistently leads, achieving $\sim$50\% higher scene coverage by step 50.

\begin{table}
    \centering
    \begin{tabular}{lccc}
        \toprule
        & \multicolumn{3}{c}{Exploration Strategy} \\
        \cmidrule(lr){2-4}
        Question Category & VLM & Closest & Ours \\
        \midrule
        Object    & \textbf{29.32} & 27.23 & 28.09 \\
        Existence & 46.41 & 39.61 & \textbf{48.01} \\
        Attribute & 42.33 & 41.43 & \textbf{42.97} \\
        Location  & 23.51 & 24.40 & \textbf{27.17} \\
        State     & 42.63 & \textbf{46.62} & 45.97 \\
        Counting  & 37.39 & 35.99 & \textbf{37.58} \\
        Knowledge & 39.66 & 40.34 & \textbf{44.83} \\
        \midrule
        \multicolumn{1}{l}{Overall} & 37.82 & 36.93 & \textbf{39.54} \\
        \bottomrule
    \end{tabular}
    \caption{Visually grounded LLM-Match for each question category in Express-Bench~\cite{EXPRESSBench} comparing different EQA exploration approaches, VLM-only, closest-frontier, and ours.
    }
    \vspace{-1 em}
\label{tab:LLM_match_scores_breakdown_ablation}
\end{table}

\begin{figure}
  \centering
   \includegraphics[width=\linewidth]{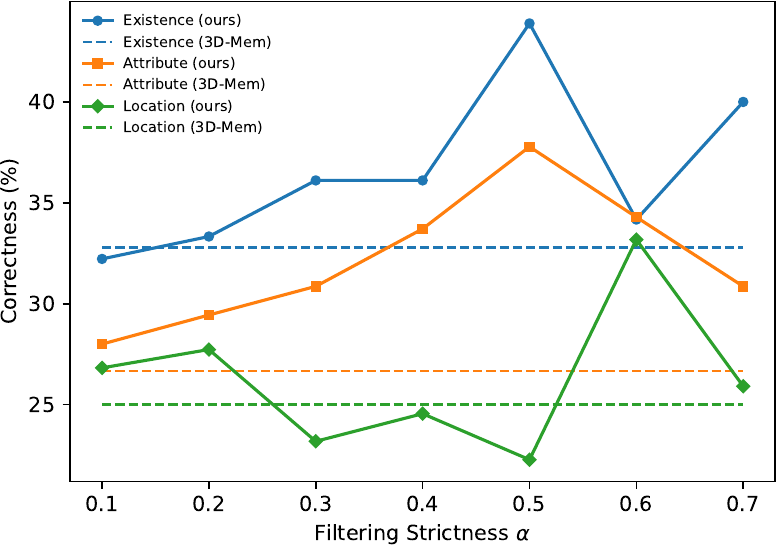}
   \caption{Curves of our question category performance vs alpha value on our Express-Bench~\cite{EXPRESSBench} tuning set. For certain categories stricter or less strict filtering improves performance over the 3D-Mem baseline.}
   \label{fig:alpha_tuning}
   \vspace{-1em}
\end{figure}

\subsection{Ablations}
\label{ablations:alpha_ablation}

\textbf{Impact of Pruning Sensitivity}. To study how the strictness of frontier pruning affects downstream performance, we evaluate different $\alpha$ values on our EXPRESS-Bench tuning set. Note that in all our evaluation we use $\alpha=0.5$ since it had the best overall performance on the tuning set. \cref{fig:alpha_tuning} shows final performance across question categories for each $\alpha$ level. We observe that location questions are particularly sensitive, peaking around $\alpha=0.6$. While our current method estimates a single empirical CDF (ECDF) for all question types, this sensitivity suggests that using category-specific distributions for p-value mapping could further improve performance on certain question types.

\noindent\textbf{Impact of Exploration Strategy}. To evaluate the effect of our Prune-Then-Plan based exploration strategy, we run ablations on EXPRESS-Bench using three methods: \textit{(i) a VLM-only planner} that selects each step purely from VLM scores over frontiers, \textit{(ii) a closest-frontier policy} that always moves to the nearest frontier, and \textit{(iii) our approach}, which applies pruning followed by a closest-frontier planner. Our method consistently outperforms the baselines, with notable gains in `existence', `location', and `knowledge questions' seen in \cref{tab:LLM_match_scores_breakdown_ablation}. While a closest-frontier strategy can suffice in small scenes with few options, it can easily trap agents in irrelevant directions in larger environments, reducing path efficiency. In contrast, our approach is both coverage-aware and semantically guided, balancing exploration with efficiency.

\noindent\textbf{Annotation Noise}. Our calibration procedure depends on annotator labels to help model the distribution of bad frontiers, to understand the impact of label quality on the final downstream performance, we perform experiments with varying levels of label noise. In total we have approximately 5,500 labeled frontiers of which roughly 3,600 are considered bad. To inject noise, we invert the labels for a random X\% of the labeled frontiers. The noisy data is used to build noisy empirical distribution functions for the Prune-Then-Plan evaluation on OpenEQA. We evaluate under varying levels of noise and report the results in \cref{tab:label_noise}. Even when we intentionally introduce noise by inverting 5\% of all frontier labels-the overall grounded LLM-Match/EAC performance remains unchanged at \textbf{34.3}. When the noise level is doubled to 10\% do we observe a modest drop in accuracy to 32.3, yet the performance still remains higher than the 3D-Mem baseline.

\begin{table}[ht]
    \centering
    \begin{tabular}{lcc}
        \hline
        Method & LLM-Match/EAC $\uparrow$ \\
        \hline
        3D-Mem & 29.2 \\
        Ours 0\% noise & \textbf{34.3} \\
        Ours 5\% noise & \textbf{34.3} \\
        Ours 10\% noise & 32.3 \\
        \hline
    \end{tabular}
    \caption{Comparison of our method to 3D-Mem on OpenEQA under varying levels of noise in calibration data labeling.}
    \label{tab:label_noise}
    \vspace{-1.0em}
\end{table}

\section{Conclusion}
\label{sec:conclusion}

We presented our Prune-Then-Plan method, a step-level calibration framework that stabilizes exploration in VLM–driven embodied question answering (EQA). By explicitly separating frontier pruning and planning, our method leverages VLMs not as direct decision-makers but as a calibration-based filter that helps identify and prune implausible frontier choices using a Holm–Bonferroni~\cite{MR538597} inspired step-down rule over calibrated frontier scores. In practice, this turns raw uncalibrated confidence values into more stable and reliable signals for frontier selection. Integrated within a standard EQA pipeline, 3D-Mem~\cite{yang20253dmem3dscenememory}, our approach improves exploration stability, scene coverage, and question-answer accuracy across OpenEQA~\cite{majumdar2023openeqa} and EXPRESS-Bench~\cite{EXPRESSBench}, with minimal system overhead. Beyond EQA, our findings suggest that step-level calibration can be a practical tool for systems where uncertainty directly influences spatial reasoning and task success.

\section*{Acknowledgments}

This work was partially supported by NSF-AI Engage Institute DRL-2112635, ARO Award W911NF2110220, and ONR Grant N00014-23-1-2356. The views contained in this article are those of the authors and not of the funding agency.

{
    \small
    \bibliographystyle{ieeenat_fullname}
    \bibliography{main}
}

\end{document}